\documentclass[journal,twoside,web]{ieeecolor}
\usepackage{generic}
\usepackage{cite}
\usepackage{amsmath,amssymb,amsfonts}
\usepackage{algorithmic}
\usepackage{graphicx}
\usepackage{algorithm,algorithmic}
\usepackage{hyperref}
\hypersetup{hidelinks=true}
\usepackage{textcomp}
\usepackage{epstopdf}
\usepackage{changes}

\def\BibTeX{{\rm B\kern-.05em{\sc i\kern-.025em b}\kern-.08em
    T\kern-.1667em\lower.7ex\hbox{E}\kern-.125emX}}
\begin{document}
\title{EEGDM: Label-Efficient EEG Representation Learning with Generative Diffusion Model}
\author{
Jia Hong Puah,
Sim Kuan Goh*,
Ziwei Zhang,
Zixuan Ye, Chow Khuen Chan,\\
Kheng Seang Lim*,
Si Lei Fong,
Kok Sin Woon, and Cuntai Guan
\thanks{
*Corresponding authors: simkuangoh@gmail.com, kslimum@gmail.com}
\thanks{J. H. Puah, S. K. Goh, Z. Zhang, Z, Ye were with the School of Artificial Intelligence and Robotics, Xiamen University Malaysia.} 
\thanks{J. H. Puah, K. S. Woon were with the School of Energy and Chemical Engineering, Xiamen University Malaysia.}
\thanks{C. K. Chan was with the Department of Biomedical Engineering, Universiti Malaya.} \thanks{K. S. Lim, S. L. Fong were with the Department of Medicine, Universiti Malaya.}
\thanks{K. S. Woon was with the Thrust of Carbon Neutrality and Climate Change, Hong Kong University of Science and Technology (Guangzhou).}
\thanks{C. Guan was with the College of Computing \& Data Science, Nanyang Technological University.}
}

\maketitle

\begin{abstract}
Electroencephalography (EEG) is a critical tool for monitoring brain activity and diagnosing neurological disorders such as epilepsy. However, learning meaningful representations from raw EEG signals remains challenging due to limited annotations, substantial inter-subject variability, and complex temporal dynamics. Recent EEG foundation models (FMs) have demonstrated promising performance through transformer-based architectures and large-scale self-supervised pretraining, yet they often incur substantial computational costs and exhibit diminishing returns with increasing model and dataset scale, limiting their practicality in clinical settings. To address these challenges, we propose EEGDM, a diffusion-based EEG representation learning framework. Specifically, EEGDM introduces a Structured State-Space Model for Diffusion Pretraining (SSMDP) that effectively captures long-range temporal dependencies through generative diffusion training on unlabeled EEG data. The representations are subsequently leveraged for downstream tasks via our {Latent Fusion Module} ({LFM}), which integrates multi-layer latent features of SSMDP. We evaluate EEGDM on three EEG benchmarks spanning EEG event classification (TUEV), seizure detection (CHB-MIT), and seizure classification (IIIC). Compared with existing state-of-the-art methods, including EEG FMs, EEGDM achieves {competitive performance across diverse tasks and datasets, exceeding existing methods in most cases} while requiring substantially fewer samples for both pretraining and downstream adaptation. These results demonstrate that diffusion-based learning can unlock more discriminative EEG representations, with direct implications for epilepsy diagnosis and management. Our source code and pretrained checkpoints are publicly available at: \url{https://github.com/jhpuah/EEGDM}.
\end{abstract}


\section{Introduction}
\label{sec:introduction}
\IEEEPARstart{E}{lectroencephalography} (EEG) provides a noninvasive technique for monitoring the brain by recording electrical signals arising from neural activity and pathological processes, with high temporal resolution. Its effectiveness has led to widespread use in brain–computer interfaces (BCIs), neuroscience research, and neurological diagnostics~\cite{yang2023biot,jiang2024large}. A key clinical application is the management and treatment of {epilepsy, which affects approximately 50 million patients worldwide~\cite{WHO_epilepsy_2024}, where EEG plays a critical role in diagnosis, characterization of epileptic activity}, and pre-surgical planning. Numerous machine learning and deep learning models have been developed to detect interictal epileptiform discharges (IEDs) and seizures from EEG recordings, which often include various other brain activities (e.g., eye movements, artifacts, and background activity)~\cite{huang2024eegdfus,goh2017automatic,goh2016multiway}. Despite showing initial promise, improving classification performance remains an active area of research for clinical adoption.

\begin{figure*}
    \centering
    \includegraphics[width=0.95\linewidth]{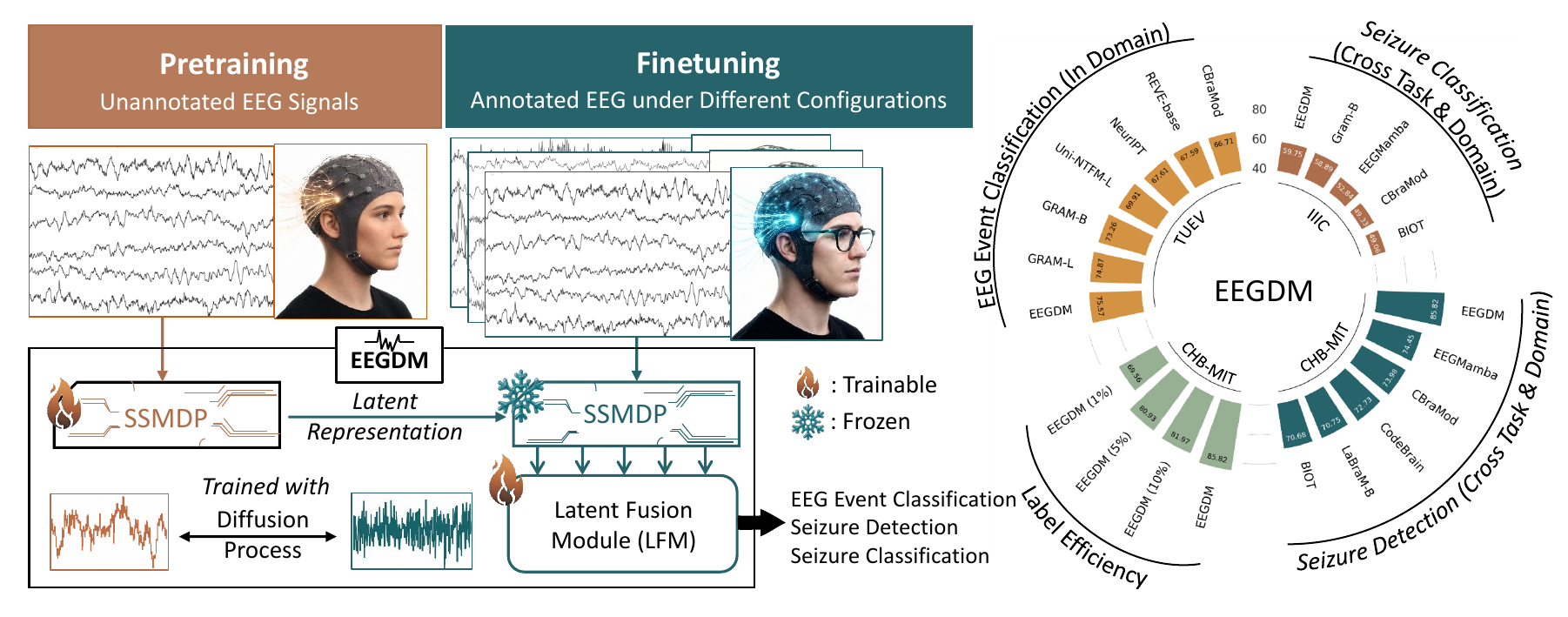}
    \caption{
    {EEGDM builds on a structured state-space model (SSMDP) learned via generative diffusion-based pretraining and finetunes a {Latent Fusion Module (LFM)} for downstream tasks. It generalizes across in-domain EEG event classification as well as cross-task and cross-domain settings, including seizure detection and seizure classification under diverse EEG configurations. {Experimental results demonstrate that EEGDM is highly competitive with recent EEG foundation models, despite being trained on substantially less data for representation learning.}
    }}
    \label{fig:ga}
\end{figure*}

A major challenge behind the performance bottleneck in IED and seizure detection is the difficulty of learning robust and meaningful representations of EEG signals, due to limited high-quality annotation and high signal variability across subjects, conditions, and brain activities~\cite{jiang2024large,wang2024eegpt}. Recently, inspired by the success of large language models (LLMs), various EEG foundation models (FMs) \cite{wang2024eegpt, jiang2025neurolm} have been proposed and achieved different degrees of success. Following the procedure of LLMs, {EEG FMs were pretrained on a large amount of unlabeled and heterogeneous EEG datasets} through self-supervised pretraining for representation learning using transformer architectures~\cite{wang2025cbramod}. In most EEG FMs,  the signals were tokenized into sequential tokens, with a subset masked to enable self-supervised training via masked token prediction~\cite{yang2023biot,jiang2024large,li2025gram}. These methods encouraged the models to learn the underlying relationships (through self- and cross-attention modules in the transformer) between tokens and effectively capture meaningful representations without relying on annotation. Subsequently, these FMs were finetuned on specific downstream tasks such as IED detection~\cite{jiang2024large}, sleep stage classification~\cite{kostas2021bendr,chin2025multi,yew2024eeg}, and emotion recognition~\cite{yi2023learning}. Although they often outperformed traditional deep learning models, they incurred high computational costs during both training and inference, with only marginal performance improvements {as model size and pretraining data scaled up. Critically, the patching and tokenization processes explicitly impose discontinuity and discretization on physically continuous EEG signals. While state-space models (e.g., Mamba and} {its variants)~\cite{gui2024eegmamba,wang2025eegmamba} have been adopted for EEG analysis to avoid patching and operate directly on continuous sequences, they have not consistently outperformed recent EEG FMs.}

Diffusion models \cite{ho2020denoising} represent another class of self-supervised and generative models that have demonstrated remarkable success in image \cite{ho2020denoising, nichol2021improved} and audio generation \cite{kongdiffwave, goel2022s}. Recent studies suggest that diffusion models also hold strong potential for representation learning \cite{xiang2023denoising}. Unlike masking-based methods, diffusion models achieve self-supervision by progressively corrupting input data with Gaussian noise in a forward process and learning to recover the original continuous data through a reverse denoising process. Several approaches \cite{baranchuklabel, mukhopadhyay2023diffusion} leverage diffusion models to learn transferable representations for visual classification and segmentation tasks. {However, the applicability and generalization of diffusion models for EEG representation learning remain largely unexplored, as most existing EEG diffusion studies focus exclusively on signal generation and data augmentation \cite{huang2025sad, li2023generative, wang2024diffmdd}. This raises a fundamental question: can diffusion models trained for continuous EEG generation serve not only as generative models but also as effective representation learners for downstream tasks?}

{To investigate this question, we propose an EEG representation learning framework, EEGDM, based on a generative diffusion model, as illustrated in Fig.~\ref{fig:ga}. Specifically, we design a structured state-space model-based diffusion pretraining architecture (SSMDP) to better capture the temporal dynamics of EEG signals and train it under a denoising diffusion probabilistic model (DDPM) framework using unannotated EEG data. The resulting latent EEG representations from the pretrained model are then leveraged for downstream classification via our proposed {Latent Fusion Module} ({LFM}).} We evaluate EEGDM on the Temple University EEG Event Corpus (TUEV) \cite{harati2015improved}, the Children’s Hospital Boston–Massachusetts Institute of Technology (CHB-MIT) dataset \cite{guttag2010chb}, {and the Ictal-Interictal-Injury Continuum (IIIC) dataset \cite{jing2023development, ge2021deep},} covering EEG event detection, seizure detection, and {seizure type classification}, respectively. We compare against state-of-the-art methods, including recent and strong EEG FMs.

\begin{itemize}
\item {We introduce \textbf{EEGDM}, a framework comprising a \textbf{Structured State-Space Model for Diffusion Pretraining (SSMDP)} that models the temporal dynamics of EEG signals for self-supervised representation learning.}
\item {We design \textbf{{Latent Fusion Module} ({LFM})} {to aggregate latent representations learned by SSMDP for robust downstream classification.}}
\item {We conduct extensive experiments on TUEV, CHB-MIT, } {and IIIC}{, demonstrating that EEGDM {is competitive with, and in most cases outperforms} state-of-the-art methods, including EEG FMs.}
\item {Through extensive experiments and ablation studies, we show that EEGDM is a strong and efficient representation learner for downstream EEG tasks.}
\end{itemize}

\section{Related Works}

This section reviews the evolution of EEG representation learning, from classical supervised, unsupervised, and semisupervised approaches to recent advances in self-supervised learning (particularly FMs and diffusion models) and their emerging potential for EEG, with an emphasis on the latter.

\subsection{Supervised and Unsupervised Approaches}

EEG representation learning has often arisen as a byproduct of supervised approaches, such as the seminal common spatial pattern methods~\cite{5593210} and more recent deep neural network models~\cite{song2022eeg}. Such representations have been examined across various neural architectures, including convolutional neural networks~\cite{8428659}, graph neural networks~\cite{tang2024multi}, and Transformers~\cite{9845479,10345766}, revealing findings that aligned with known brain activity patterns. When annotations are unavailable, unsupervised approaches, particularly autoencoders and their variants, offered an alternative. Variational autoencoder (VAE) has shown effectiveness in learning robust representation in the regularized latent space, achieving transfer learning from motor execution signal to motor imaginary signals \cite{lee2022motor}. Another approach was metric learning, which embedded EEG signals into a vector space where similarity was defined as proximity between samples, typically optimized using contrastive loss (e.g., \cite{li2025neuron}). However, these methods are constrained by the availability of annotation and often yield task-specific representations that lack generalizability.

Another line of research was semi-supervised learning that leverages a small amount of annotated data to train a model, while using the model itself to generate “pseudo-labels” for unannotated data, which can then be used as additional training data. For example, in \cite{wang2023eeg}, the empirical probabilities of the training data are used as pseudo-labels to calibrate model certainty and confidence, thereby addressing the scarcity of annotations in motor imagery classification. Another promising approach was self-supervised learning, which learns by corrupting and subsequently restoring the training data. Common corruption strategies included masking, adding noise, or combining both. These techniques have been successfully applied to various EEG tasks, such as sleep stage classification \cite{chien2022maeeg}, evoked response enhancement \cite{chuang2022convolutional}, {emotion recognition \cite{gong2023astdf},} and motion intention recognition \cite{zhang2025dmae}. Recently, self-supervised pretraining methods, particularly mask prediction with transformers and diffusion models, have demonstrated strong potential across multiple domains. However, only a few pioneering studies have begun to explore their effectiveness in EEG representation learning.

\subsection{EEG FMs}
{Inspired by LLMs, EEG FMs employ self-supervised pretraining strategies, including mask prediction~\cite{jiang2024large}{, contrastive learning~\cite{chen2025large} and anchor alignment \cite{shao2026dare}}, on large-scale heterogeneous EEG data} using transformer architectures, followed by finetuning on labeled data in a supervised manner. The pretraining allowed EEG FMs to grasp the underlying structure of EEG from various problem domains and experimental paradigms, thus addressing the annotation scarcity issue and reducing the risk of overfitting \cite{rafiei2022self}. Prior to pretraining, the raw EEG signal was usually preprocessed and then segmented into patches. A tokenizer, usually a variant of vector-quantized (VQ) VAE \cite{van2017neural}, was trained to encode these patches into discrete representations, known as tokens. The VQ encoder first projected a patch to a vector representation, then the representation was replaced by the most similar entry (token) from a codebook. The VQ decoder then reconstructed a target from the token, either the original patch itself (in CBraMod \cite{wang2025cbramod} and Gram \cite{li2025gram}) or a transformation of the original patch, such as Fourier spectrum (in LaBraM \cite{jiang2024large}). This process ensured that the tokens in the codebook capture semantically meaningful information about the original patches.

\begin{figure*}
    \centering
    \includegraphics[width=0.95\linewidth]{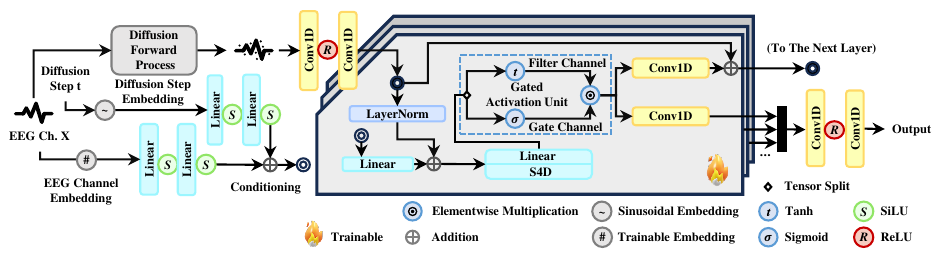}
    \caption{
    Illustration of the proposed SSMDP for representation learning. The model learned to denoise and reverse a noised EEG signal by predicting the diffusion velocity, conditioned on diffusion step embedding and channel embedding. SSMDP built upon DiffWave and S4D architectures, and trained using DDPM framework. Subsequently, the latent activities from the gate or filter channels were then leveraged for downstream classification.
    }
    \label{fig:backbone-arch}
\end{figure*}

Subsequently, self-supervised pretraining operates on the sequence tokenized by the VQ encoder and the codebook. Initially, a subset of a sequence is masked by randomly replacing some of the tokens with a ``mask token''. The masked sequence is then processed by a transformer-based model to predict the index of the replaced tokens in the codebook. The neural architecture of the model varies: LaBraM \cite{jiang2024large} used a classic transformer, CBraMod \cite{wang2025cbramod} employed a criss-cross attention mechanism to model spatial and temporal dependencies in parallel, and Gram \cite{li2025gram} used a masked autoencoder consisting of a layer-fusion encoder and a transformer-based decoder. Subsequently, the pretrained model can be finetuned for classification tasks. Usually, a trainable ``class token'' is concatenated into the input sequence to be processed by transformer. Then, a linear classification head was trained to classify the processed class token to a label. While EEG FMs often outperformed classical deep learning models, they incurred high computational costs during both training and inference, with only marginal performance improvements as model {and data} size increases~\cite{fang2026neuript,chen2025uni}.

\subsection{Diffusion Model in Representation Learning}
Diffusion models are closely related to denoising autoencoders \cite{bengio2013generalized}, which are capable of producing meaningful representations in their latent space. In particular, denoising diffusion autoencoder \cite{xiang2023denoising} showed that the inner working of diffusion models was analogous to an autoencoder. Several works explored the efficiency of latent activities of diffusion models in image classification \cite{mukhopadhyay2023diffusion} and image segmentation task\cite{baranchuklabel}.

In addition to latent activities, the connection between conditional diffusion models and conditional probability distributions also led to alternative ways in representation learning. Diffusion classifier \cite{li2023your} showed that any conditional diffusion model can be converted into a classifier without additional training. Joint diffusion model \cite{deja2023learning} unified this generator-classifier duality: the model was explicitly trained to minimize both classification loss and generative loss, so it can perform both tasks in a single model. These findings presented a promising direction for generalization to EEG representation learning and classification tasks, despite being underexplored.

\subsection{Diffusion Models for EEG Signals}

There is limited work on adapting diffusion models to EEG signal representation learning. Early works have primarily used EEG signals as conditioning inputs to generate other modalities, such as images \cite{chen2025exploring} and audio \cite{qi2023audiodiffusion}. Inspired by image super-resolution (e.g., \cite{moser2024diffusion}), \cite{wang2025generative} proposed a diffusion model that can generate high-resolution EEG from low-resolution EEG to mitigate the data scarcity issue for downstream classification tasks and improve spatial resolution for signal source localization. In \cite{huang2024eegdfus}, a conditional diffusion model was applied to generate a clean EEG signal under the guidance of noisy EEG signals, hence suppressing EEG artifacts such as electrooculography artifacts.

Several works \cite{huang2025sad, li2023generative, wang2024diffmdd} quantitatively evaluated the similarity between the generated and real signals, and used the synthesized data for augmentation in various downstream tasks, such as imagined motor activity detection and major depressive disorder classification. Nonetheless, the applicability and generalization of diffusion models as EEG representation learners remain underexplored.

\begin{figure*}
    \centering
    \includegraphics[width=0.95\linewidth]{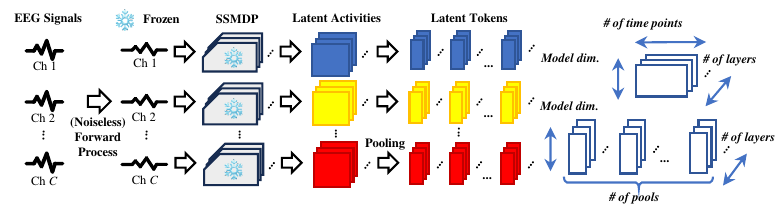}
    \caption{
    From the pretrained SSMDP, we extract the latent activities of each EEG channel as a four-dimensional tensor: (\# channels, \# layers, \# time points, \# gate channels per layer). The latent activities are subsequently pooled along the temporal dimension to form compact latent tokens, reducing the representation to (\# channels, \# layers, \# pools, \# gate channels per layer). Different colors denote different EEG channels.
    }
    \label{fig:ext-and-pool}
\end{figure*}

\section{Methodology}
We begin by introducing the key components upon which EEGDM is built, namely diffusion models and structured state-space models, before presenting the EEGDM framework.

\subsection{Preliminaries}

\subsubsection{Diffusion Model}
Diffusion models consist of a forward process and a reverse process. The forward process gradually transforms a target data distribution into a standard Gaussian distribution, while the reverse process learns to denoise step by step to reconstruct the original data. After training, the reverse process can be used for generation by gradually transforming a sample from the Gaussian distribution into one that resembles the target data distribution.

The forward process models the trajectory of a data point $\mathbf{x}_0$ drifts and diffuses from its original distribution $p(\mathbf{x}_0)$ to the Gaussian distribution $\mathcal{N}(0,I)$. This process is split into $T$ diffusion steps, where each step is an intermediate distribution dependent on the original data and a noise schedule $\bar\alpha$: 
\begin{equation}
    p(\mathbf{x}_t|\mathbf{x}_0) = \sqrt{\bar\alpha_t} \mathbf{x}_0 + \sqrt{1 - \bar\alpha_t} \epsilon
    \label{eq:diff-fwd}
\end{equation}
where $t=1,2,3\dots,T$, $\epsilon\sim\mathcal{N}(0,I)$, and $\bar\alpha_t\in(0,1)$ is a set of strictly decreasing hyperparameters controlling the noise schedule.

The reverse process starts from a pure Gaussian distribution $q(\mathbf{x}_T)$. The model is trained to perform denoising under the assumption that each transition step in the reverse process is also Gaussian. This assumption becomes increasingly valid as the number of diffusion steps $T$ grows large. The reverse process is modeled as:
\begin{equation}
    q_{\theta}(\mathbf{x}_{t-1} | \mathbf{x}_t) = \mathcal{N}(
        \mathbf{x}_{t-1};
        \mu(\mathbf{x}_t, t),
        \sigma_t^2
    )
\end{equation}
parameterized by mean $\mu$ and variance $\sigma_t^2$. $\sigma_t^2$ is assumed to be a function of $\bar\alpha_t$ and not dependent on $\mathbf{x}_t$. Previous works have found that, instead of predicting $\mu$, predicting added noise proposed in DDPM framework \cite{ho2020denoising} and velocity \cite{salimansprogressive}:
\begin{equation}
    v_t=\sqrt{\bar\alpha_t} \epsilon - \sqrt{1 - \bar\alpha_t} \mathbf{x}_0
    \label{eq:diff-velocity}
\end{equation}
give better results in general. 

A deep neural network $\Phi$ is trained to approximate the chosen target $v_t$ with a training loss $\mathcal{L}$ to maximize the evidence lower bound of $q_{\theta}(\mathbf{x}_0)$, which is typically simplified to minimize the mean square error (MSE) between the prediction and the actual target of every diffusion step \cite{ho2020denoising}:
\begin{equation}
    \mathcal{L_{\text{MSE}}}=\mathbb{E}_{t\sim[1,T]} \left[\lVert v_t-\Phi(\mathbf{x}_t,t) \rVert^2 \right]
    \label{eq:diff-sim-loss}
\end{equation}

Diffusion models do not have restrictions on the architecture of the backbone model $\Phi$, some popular choices include U-Net \cite{ronneberger2015u} and transformer \cite{vaswani2017attention}. Conditional diffusion models accept additional inputs $c$ known as ``conditioning'' that provide supplementary information about the target $\mathbf{x}_0$, such as its class label. This effectively converts the outcome of the reverse diffusion process into conditional distributions $q_{\theta}(\mathbf{x}_0|c)$, allowing for a finer control over the generation outcome \cite{ho2022classifier}.

\subsubsection{Structured State-Space Model}
Structured state-space models \cite{gu2021efficiently} (SSMs) are a class of neural networks that originate from the state-space representation of a dynamic system in control theory. SSMs project a continuous-time input $x(\tau)$ to a continuous-time output $y(\tau)$ and a hidden state $h(\tau)$ with matrices $\mathbf{A},\mathbf{B},\mathbf{C}$ and $\mathbf{D}$:
\begin{subequations}
\label{eq:ssm-cont}
\begin{align}
    h'(\tau) &= \mathbf{A}h(\tau) + \mathbf{B}x(\tau)
    \label{eq:ssm-cont-1}
    \\ 
    y(\tau) &= \mathbf{C}h(\tau) + \mathbf{D}x(\tau)
    \label{eq:ssm-cont-2}
\end{align}
\end{subequations}

In practice, continuous-time data are often unavailable, so SSMs are typically computed using their discrete-time approximations:
\begin{subequations}
\label{eq:ssm-disc}
\begin{align}
    h_k &= \overline{\mathbf{A}}h_{k-1} + \overline{\mathbf{B}}x_k 
    \label{eq:ssm-disc-1}
    \\
    y_k &= \mathbf{C}h_k + \mathbf{D}x_k
    \label{eq:ssm-disc-2}
\end{align}
\end{subequations}

The matrices in \eqref{eq:ssm-disc-1} can be trained directly or obtained by applying a discretization rule with trainable step size $\Delta$ on \eqref{eq:ssm-cont-1}, like zero-order hold:
\begin{equation}
    \overline{\mathbf{A}} = \exp(\Delta\mathbf{A}),\text{ }
    \overline{\mathbf{B}} = (\Delta\mathbf{A})^{-1} ({\exp(\Delta\mathbf{A}) - I})(\Delta\mathbf{B})
\end{equation}
This approach preserves the continuous-time structure of SSMs, enabling the model to operate at different sampling rates without retraining through the adjustment of $\Delta$ \cite{gu2021efficiently}. However, in practice, this advantage may be diminished if the SSM is used alongside components that are inherently dependent on the sampling rate.

Unrolling the formulae in \eqref{eq:ssm-disc} will eliminate $h_k$, which results in the convolution form of SSMs that can be trained efficiently:
\begin{equation} 
\begin{split}
    \overline{K} &=(\overline{\mathbf{CB}},\overline{\mathbf{CAB}},\overline{\mathbf{CA}}^2\overline{\mathbf{B}},\dots,\overline{\mathbf{CA}}^{L-1}\overline{\mathbf{B}}) \\
    y &= \overline{K} \ast x = \sum_{i=0}^{k}\overline{\mathbf{CA}}^{i}\overline{\mathbf{B}}x_{k - i},\quad k = 0,\dots,L-1
\end{split}
\label{eq:ssm-conv}
\end{equation}

However, in practice, the calculation of $\overline{\mathbf{A}}^i$ in \eqref{eq:ssm-conv} is a performance bottleneck and is vulnerable to the vanishing gradient problem. The S4D \cite{gu2022parameterization} overcomes this by constraining $\mathbf{A}$ to be a diagonal matrix with special initialization schemes.

\begin{figure*}
    \centering
    \includegraphics[width=0.95\linewidth]{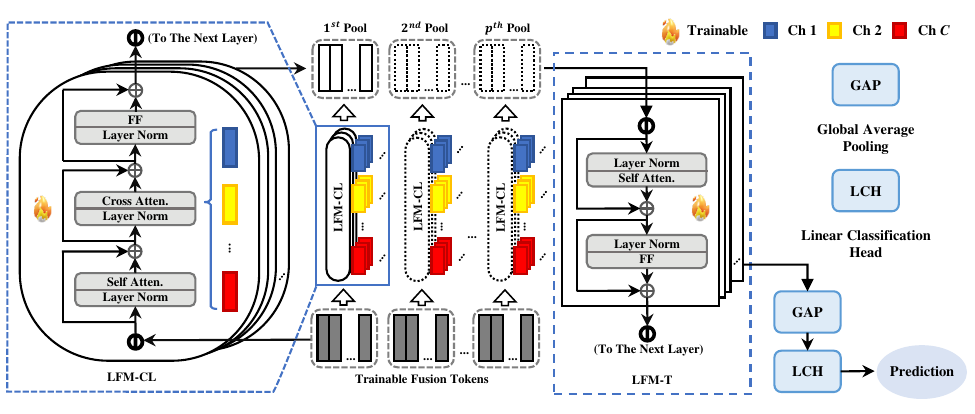}
    \caption{Architecture of the proposed {LFM}. {LFM} comprises two main modules: {LFM-CL} and {LFM-T}. \textbf{Left:} The LFM-CL processes tokens from one pool at a time, with each block handling tokens from a single layer. Trainable fusion tokens are passed sequentially between blocks, integrating nonisomorphic representations from different layers via cross-attention. \textbf{Middle:} To capture time-invariant representations, the same fusion model is applied to every pool. \textbf{Right:} The {LFM-T} {processes the} fused representations from all pools. Subsequently, {global average pooling aggregates the learned representations, followed by a linear classification head that produces the final predictions.}}
    \label{fig:decoder-arch}
\end{figure*}

\subsection{EEGDM}

Building upon the aforementioned components, EEGDM comprises two main modules: (1) SSM-Based Diffusion Pretraining (SSMDP) for EEG representation learning, and (2) {LFM} for downstream classification tasks. We detail the neural architectures and training procedures of each component in the following subsections.

\subsubsection{SSM-Based Diffusion Pretraining}

{The backbone of SSMDP is built upon DiffWave~\cite{kongdiffwave}, a well-established diffusion backbone for unconditional audio generation. We further replace and enhance the original dilated convolutional layers with S4D~\cite{gu2022parameterization} architectures, enabling efficient modeling of long-range temporal dependencies. In addition, we introduce channel-aware conditioning to adapt the diffusion process for} {multichannel EEG generation.} The architecture, illustrated in Fig.~\ref{fig:backbone-arch}, is designed to learn robust representations of EEG signals by preserving fine-grained temporal structures while capturing long-range dependencies across extended time horizons. One limitation of S4D is its unidirectional nature: it models each output $y_k$ as a function of current and past inputs $x_{0:k}$, while completely ignoring future inputs $x_{k+1:L}$. To model bidirectional dependencies, we adopt a bidirectional-SSM setup, wherein one S4D processes the original sequence in the forward direction and another processes the reversed sequence, as suggested in previous work \cite{goel2022s}. 

SSMDP processed signals from each EEG channel individually, and 128-dimensional trainable channel embeddings were included to encode the spatial context of input signals, while the diffusion step was embedded through sinusoidal embedding \cite{vaswani2017attention}. The sum of these two embeddings is fed to the backbone model as conditioning, informing the model about the differences in spatial coordinate and noise level. Subsequently, SSMDP was trained using the DDPM framework. EEG signals were first subjected to a forward diffusion process, progressively injecting noise into the signals. Our model then learned to reverse this process step by step, denoising the input and learning meaningful latent representations.

Our SSMDP consisted of twenty layers of SSMDP blocks, each block contained 128 gate channels and 128 filter channels. Each channel, either gate or filter, had its own trainable bidirectional S4D with 128-dimensional hidden state. The S4Ds were initialized with diag-lin scheme (see \cite{gu2022parameterization}).

\subsubsection{{Latent Fusion Module}}

{Our LFM integrates latent activities derived from SSMDP. To reduce the computational overhead associated with high-dimensional latent representations, we apply window-based mean and standard deviation pooling to aggregate the latent activities into compact latent tokens, as illustrated in Fig. \ref{fig:ext-and-pool}. LFM comprises LFM-CL and {LFM-T}, illustrated in Fig. \ref{fig:decoder-arch}. LFM-CL integrates latent activities into latent fusion tokens from the gate or filter channels of the gated activation unit of the SSMDP {across EEG \textbf{C}hannels and SSMDP \textbf{L}ayers}, illustrated in Fig. \ref{fig:decoder-arch}. Gate channels were selected based on empirical analysis (see Section~\ref{sec:latentSSMDP}).} 
{The fused representations were then passed to the LFM-T, which captured dependencies across \textbf{T}emporal pools, and followed by a linear classification head with softmax activation.}

{Our LFM has twenty LFM-CL blocks and eight LFM-T blocks, each with 8-head attention. The attention embedding dimension and MLP hidden dimension are set to 128 and 512, respectively.}

\subsubsection{Training Procedures}

\paragraph{Signal Embedding}\label{para:signal-embed}

EEG data exhibit a wide dynamic range with occasional extreme values, which can hinder model training. To address this, we apply the $\mu$-law companding:
\begin{equation}
    u(x)=
    \begin{cases}
       \text{sgn}(x)\frac{\ln(1+\mu|x|)}{\ln (1+\mu)}, &\quad\text{if }|x| > 1\\
        x, &\quad\text{otherwise.}
     \end{cases}
\end{equation}
where $x$ denotes the normalized signal and $\mu=255$. This nonlinear transformation compressed the large-amplitude values more than the smaller ones, effectively suppressing outliers and enhancing numerical stability. In addition to signal embedding, we included learnable embeddings for EEG channels and sinusoidal embeddings for diffusion timesteps to provide spatial and noise-level context to the model, as shown in Fig.~\ref{fig:backbone-arch}.

\paragraph{Pretraining of SSMDP}\label{para:noise-sch}

We adopted DDPM and modified the noise schedule as EEG signals are approximately Gaussian-distributed \cite{chen2014electroencephalogram} and typically exhibit low signal-to-noise ratios. To preserve signal integrity across diffusion steps, we adopt a less aggressive noise injection scheme compared to standard DDPM configurations. Specifically, the total diffusion timestep $T$ was set to 50 (in contrast to the typical 1000), and we employed the cosine noise schedule \cite{nichol2021improved}.

\begin{table}[!t]
\setlength{\tabcolsep}{2pt}
\caption{{Comparative analysis (\%) on Multi-event TUEV with FMs and non-FMs.}}
\label{table:main-res}
\centering
\begin{tabular}{l r r r r}
\hline
\textbf{Model} & \multicolumn{1}{c}{\textbf{Size}} & \multicolumn{1}{c}{\textbf{Kappa}} & \multicolumn{1}{c}{\textbf{BAcc}} & \multicolumn{1}{c}{\textbf{WF1}} \\
\hline
\multicolumn{5}{c}{\textbf{\textit{Training from Scratch (non-FM)}}}\\

SPaRCNet \cite{jing2023development}     & 0.79M     &42.33 ± 1.81 &41.61 ± 2.62 &70.24 ± 1.04\\
ContraWR \cite{yang2021self}            & 1.6M      &39.12 ± 2.37 &43.84 ± 3.49 &68.93 ± 1.36\\
FFCL \cite{li2022motor}                 & 2.4M      &37.32 ± 1.88 &39.79 ± 1.04 &67.83 ± 1.20\\
CNN-Trans \cite{peh2022transformer}     & 3.2M      &38.15 ± 1.34 &40.87 ± 1.61 &68.54 ± 2.93\\
ST-Trans \cite{song2021transformer}     & 3.5M      &37.65 ± 3.06 &39.84 ± 2.28 &68.23 ± 1.90\\
\hline
\multicolumn{5}{c}{\textbf{\textit{Finetuning of Pretrained Parameters (FM)}}}\\

BIOT \cite{yang2023biot}                & 3.2M      &52.73 ± 2.49 &52.81 ± 2.25 &74.92 ± 0.82\\
LaBraM-Base \cite{jiang2024large}       & 5.8M      &64.33 ± 0.87 &67.25 ± 1.02 &82.01 ± 0.87\\
LaBraM-Large \cite{jiang2024large}      & 46M       &66.22 ± 1.36 &65.81 ± 1.56 &83.15 ± 0.40\\
LaBraM-Huge \cite{jiang2024large}       & 369M      &66.80 ± 1.85 &66.95 ± 1.54 &84.01 ± 0.38\\
CBraMod \cite{wang2025cbramod}          & 4.0M      &67.72 ± 0.96 &66.71 ± 1.07 &83.42 ± 0.64\\
Gram-B \cite{li2025gram}                & 6.0M      &65.28 ± 1.79 &73.26 ± 0.93 &86.14 ± 0.80\\
Gram-M \cite{li2025gram}                & 47.19M    &66.77 ± 2.27 &74.06 ± 1.26 &86.74 ± 1.17\\
Gram-L \cite{li2025gram}                & 251.28M   &\underline{71.30 ± 3.33} &\underline{74.87 ± 1.51} &\textbf{88.24 ± 1.58}\\

{REVE-Base~\cite{el2026reve}} & {69M} & {67.83 ± 1.99} & {67.59 ± 2.29} & {84.51 ± 1.29} \\
{NeurIPT~\cite{fang2026neuript}} & {73.5M} & {69.70 ± 1.85} & {67.61 ± 1.33} & {84.28 ± 0.89} \\
{CodeBrain~\cite{ma2025codebrain}} & {15.17M} & {69.12 ± 1.01} & {64.28 ± 0.62} & {83.62 ± 0.48}\\
{Uni-NTFM-L~\cite{chen2025uni}} & {1900M} & {70.30 ± 1.48} & {69.91 ± 1.70} & {84.66 ± 1.32} \\

\hline
\multicolumn{5}{c}{\textbf{\textit{Freezing of Pretrained Parameters (FM)}}}\\

MMM \cite{yi2023learning}               & 0.36M     &23.22 ± 1.55 &52.69 ± 0.71 &68.20 ± 1.93\\
BENDR \cite{kostas2021bendr}            & 157.14M   &39.64 ± 1.48 &50.09 ± 2.24 &75.35 ± 1.07\\
EEGPT-Tiny \cite{wang2024eegpt}         & 4.7M      &50.85 ± 1.73 &56.70 ± 0.66 &75.35 ± 0.97\\
EEGPT \cite{wang2024eegpt}              & 25M       &63.51 ± 1.34 &62.32 ± 1.14 &81.87 ± 0.63\\
NeuroLM-B\cite{jiang2025neurolm}        & 254M      &42.85 ± 0.48 &45.60 ± 0.48 &71.53 ± 0.28\\
NeuroLM-L\cite{jiang2025neurolm}        & 500M      &44.14 ± 9.96 &41.32 ± 12.35 &73.87 ± 4.00\\
NeuroLM-XL\cite{jiang2025neurolm}       & 1696M     &45.70 ± 4.98 &46.79 ± 3.56 &73.59 ± 2.19\\
\hline
\multicolumn{5}{c}{\textbf{\textit{Freezing of Pretrained Parameters (non-FM)}}}\\
\textbf{EEGDM (Ours)} & 12.8M & \textbf{74.23 ± 1.36} & \textbf{75.57 ± 1.47} & \underline{86.88 ± 0.66}\\
\hline
\end{tabular}

\end{table}

\begin{table}[!t]
\setlength{\tabcolsep}{3pt}
\caption{Ablations of $\mu$-law, noise scheduler, noiseless forward process, {and bidirectional SSM} (\%).}
\label{table:alb-misc}
\centering
\begin{tabular}{l r r r}
\hline
\textbf{Model Setting} & \multicolumn{1}{c}{\textbf{Kappa}} & \multicolumn{1}{c}{\textbf{BAcc}} & \multicolumn{1}{c}{\textbf{WF1}} \\
\hline
\textbf{EEGDM} & \textbf{74.23 ± 1.36} & \textbf{75.57 ± 1.47} & \textbf{86.88 ± 0.66}\\
\ - $\mu$-law Signal Embedding & \underline{72.50 ± 3.02} & {69.12 ± 1.81} & \underline{85.79 ± 1.41}\\
\ - Cosine Noise Schedule & {70.97 ± 1.04} & \underline{71.19 ± 0.92} & {85.07 ± 0.49}\\
\ - Noiseless Forward Process & {71.39 ± 1.46} & {56.19 ± 3.34} & {84.87 ± 0.54}\\
\ {- Bidirectional SSM} & {65.92 ± 0.95} & {65.57 ± 2.01} & {82.31 ± 1.07}\\
\hline
\end{tabular}
\end{table}

\paragraph{Finetuning of {LFM}}
During finetuning, the \textit{latent activities} from all gate channels across all SSMDP layers were collected. As illustrated in Fig.~\ref{fig:ext-and-pool}, given an EEG segment with $C$ channels and $L$ samples, processed by a backbone model with $n$ layers and $H$ gate channels, the resulting latent activities form a tensor of shape $(C, n, L, H)$. We divided the time axis ($L$) into $p$ segments and performed pooling, yielding latent tokens with shape $(C, n, p, H)$. The gate dimension $H$ serves as the embedding dimension, so each token summarizes latent activity within a local window of $L/p$ samples for a specific channel and layer.

Latent fusion was performed {by applying the LFM-CL (Fig.~\ref{fig:decoder-arch}, left) independently to each pooled representation.} For each temporal pool, there were $n$ groups of $C$ latent tokens (each of dimension $H$) that represented activity across EEG channels. Since each group encoded different information, we assign a dedicated decoder block to each group. $N$ trainable latent fusion tokens were sequentially passed through these decoder blocks: first through the block for the first group, then the second, and so on. Within each decoder block, the fusion tokens acted as queries, and the corresponding latent tokens served as keys and values. This design condensed the $C$-channel information into $N$ context-rich fusion tokens for each temporal pool. In this work, $n=20$, $L=1000$, $H=128$, and $N=16$. $C=22$ for TUEV and $16$ for CHB-MIT and IIIC. $p$ was the number of seconds in the EEG segment, which is 5 for TUEV and 10 for CHB-MIT and IIIC.

The resulting fusion tokens from all $p$ temporal pools were concatenated and passed to {the {LFM-T} (Fig.~\ref{fig:decoder-arch}, right)}. This {module} captured dependencies across pools and fusion tokens. Then, its outputs were global average-pooled and fed into a linear classification head. The {{LFM}} was trained in a supervised manner, representing the finetuning phase of the EEGDM framework.

\section{Experiment}
The experimental details include: (1) dataset preprocessing, (2) training configurations and hyperparameters, and (3) evaluation and baseline comparisons.

\subsection{Dataset and Preprocessing}
The TUEV \cite{harati2015improved} dataset is a multi-event dataset from the TUH EEG Corpus \cite{obeid2016temple} for the detection and classification of epileptic events from EEG recordings, including other brain activities such as eye movements, artifacts, and background. The recordings were sampled at a 250 Hz sampling rate and referenced using a transverse central parietal (TCP) montage. The recordings in the dataset were labeled with the following annotations: spike and slow wave (SPSW), generalized periodic epileptiform discharge (GPED), periodic lateralized epileptiform discharge (PLED), eye movement (EYEM), artifact (ARTF), and background (BCKG).

The CHB-MIT \cite{guttag2010chb} dataset was collected at Children’s Hospital Boston from pediatric patients with intractable seizures. It contains EEG recordings from 24 subjects, annotated with binary labels: seizure and non-seizure. The recordings were sampled at 256 Hz using the International 10–20 system. {Due to the severe class imbalance in CHB-MIT approximately 1:150, class weighting was applied during downstream fine-tuning. The seizure class weight was selected through a coarse grid search over {50,100,150} using the validation set, with 100 chosen. The variations among different weighting schemes resulted in only marginal performance differences.
}

{The IIIC dataset \cite{jing2023development, ge2021deep} is a collection of EEG signals that contain seizures and seizure-like periodic activities, including 
seizure (SZ), lateralized periodic discharges (LPD), generalized periodic discharges (GPD), lateralized rhythmic delta activity (LRDA), generalized rhythmic delta activity (GRDA), and ``Other''. The recordings were sampled at 200 Hz sampling rate using the International 10-20 system.}

{For TUEV and CHB-MIT datasets, we followed the same preprocessing steps and data splitting procedure as CBraMod \cite{wang2025cbramod}, including band-pass filtering (0.1--75 Hz) with a 50 Hz notch filter, resampling to 200 Hz, and amplitude normalization (scaling by 1/100 to ensure values are within approximately [-1, 1]). For TUEV, recordings were segmented into 5-second epochs, and the training set was further} {divided into 80\% training and 20\% validation splits. For CHB-MIT, recordings were segmented into 10-second epochs. Subjects 1--20 were used for training, subjects 21--22 for validation, and subjects 23--24 for testing. }

{For the IIIC dataset~\cite{jing2023development, ge2021deep}, we evaluated cross-subject generalization using a subject-wise 60:20:20 split. The exact split configuration is publicly available in our GitHub repository to ensure reproducibility. The signal preprocessing steps (filtering, resampling and normalization) are consistent with the other two datasets. }

\begin{table}[!t]
\caption{Impacts of different diffusion time step (\%).}
\label{table:alb-diff-step}
\centering
\begin{tabular}{l r r r}
\hline
\textbf{Step} & \multicolumn{1}{c}{\textbf{Kappa}} & \multicolumn{1}{c}{\textbf{BAcc}} & \multicolumn{1}{c}{\textbf{WF1}} \\
\hline

\textbf{No} & \underline{74.23 ± 1.36} & \textbf{75.57 ± 1.47} & \underline{86.88 ± 0.66}\\
1 & \textbf{74.46 ± 1.39} & \underline{74.87 ± 1.83} & \textbf{86.94 ± 0.68}\\
2 & {73.73 ± 1.16} & {74.83 ± 1.55} & {86.66 ± 0.55}\\
3 & {69.05 ± 1.63} & {69.89 ± 1.49} & {84.34 ± 0.81}\\
\hline
\end{tabular}
\end{table}

\begin{table}[!t]
\caption{Impacts of pooling strategies and latent activities (\%).}
\label{table:alb-latent-type-pool}
\centering
\begin{tabular}{l l r r r}
\hline
\textbf{Pooling} & \textbf{Latent} & \multicolumn{1}{c}{\textbf{Kappa}} & \multicolumn{1}{c}{\textbf{BAcc}} & \multicolumn{1}{c}{\textbf{WF1}} \\
\hline

\textbf{Std} & \textbf{Gate} & \textbf{74.23 ± 1.36} & \textbf{75.57 ± 1.47} & \textbf{86.88 ± 0.66}\\

Std & Filter & \underline{73.46 ± 2.33} & \underline{68.74 ± 0.63} & \underline{86.49 ± 1.18}\\
Average & Gate &  {71.66 ± 1.13} & {57.71 ± 1.89} & {85.24 ± 0.57}\\
Average & Filter & {69.83 ± 1.92} & {68.33 ± 3.17} & {84.46 ± 1.05}\\

\hline
\end{tabular}
\end{table}

\begin{table}[!t]
\setlength{\tabcolsep}{5pt}
\caption{Impacts of using latent activities from different layers (\%).}
\label{table:alb-half}
\centering
\begin{tabular}{l r r r r}
\hline
\textbf{Layers} & \multicolumn{1}{c}{\textbf{Size}} & \multicolumn{1}{c}{\textbf{Kappa}} & \multicolumn{1}{c}{\textbf{BAcc}} & \multicolumn{1}{c}{\textbf{WF1}} \\
\hline
\textbf{All} & 6.9M & \textbf{74.23 ± 1.36} & \textbf{75.57 ± 1.47} & \textbf{86.88 ± 0.66}\\
\hline
First half & 7.0M & {62.97 ± 2.15} & {64.57 ± 1.54} & {81.00 ± 1.15}\\
Second half & 7.0M &  \underline{71.60 ± 1.14} & \underline{68.40 ± 1.92} & \underline{85.60 ± 0.52}\\
\hline
First quarter & 6.9M &  {59.63 ± 2.49} & {58.75 ± 3.13} & {78.88 ± 1.43}\\
Second quarter & 6.9M &  {68.40 ± 2.66} & \underline{69.47 ± 1.45} & {83.80 ± 1.35}\\
Third quarter & 6.9M &  {68.32 ± 1.22} & {64.39 ± 1.37} & {83.96 ± 0.63}\\
Fourth quarter & 6.9M &  \underline{70.48 ± 2.21} & {65.78 ± 1.46} & \underline{84.71 ± 1.16}\\
\hline

\end{tabular}
\end{table}

\begin{table}[!t]
\caption{Impacts of different latent fusion strategies (\%).}
\label{table:alb-fusion}
\centering
\begin{tabular}{l r r r r}
\hline
\textbf{Fusion} & \multicolumn{1}{c}{\textbf{Size}} & \multicolumn{1}{c}{\textbf{Kappa}} & \multicolumn{1}{c}{\textbf{BAcc}} & \multicolumn{1}{c}{\textbf{WF1}} \\
\hline
\textbf{Base} & 6.9M & \textbf{74.23 ± 1.36} & \textbf{75.57 ± 1.47} & \textbf{86.88 ± 0.66}\\
No & 7.0M & {61.60 ± 3.69} & {51.13 ± 7.15} & {79.93 ± 2.26}\\
Mean & 7.0M &  \underline{65.05 ± 1.41} & \underline{59.40 ± 8.67} & \underline{82.08 ± 0.83}\\

\hline
\end{tabular}
\end{table}

\subsection{Training Configurations and Hyperparameters}
The SSMDP was trained with a batch size of 108 for 100 epochs using the AdamW optimizer. The prediction target was velocity (see \eqref{eq:diff-velocity}). The learning rate, weight decay, $\beta_1$, and $\beta_2$ were $10^{-4}$, 0, 0.99, and 0.999, respectively. An exponential moving average (EMA) was applied to the model with a decay rate of 0.999. Gradients were clipped by their norm with a threshold of 1. The model size of SSMDP was 5.9M parameters.

{Unless otherwise specified, the LFM was trained with AdamW optimization without class weighting and with a label smoothing factor of 0.1.} A batch size of 256 was used, evenly distributed across GPUs. The initial learning rate was set to $10^{-5}$ and scheduled using a one-cycle cosine annealing policy, with a peak learning rate of $5\times10^{-4}$ and a warm-up period of 5 epochs. To stabilize training, an exponential moving average was applied with a decay rate of 0.999. Additionally, gradients were clipped by their global norm, with a threshold of 3. The model size of LFM was 6.9M parameters.

The LFM was trained for a maximum of 50 epochs, with early stopping applied when the validation Cohen’s kappa score did not improve for 3 consecutive epochs after epoch 20. The checkpoint achieving the highest validation score was subsequently selected for evaluation on the test set.

\subsection{Evaluation and Baselines}
The classification performance was evaluated using balanced accuracy (``BAcc''), Cohen's kappa (``kappa''), and weighted F1 score (``WF1''). We reported the mean and standard deviation of performance across five independent runs. The selected baselines include deep learning models trained from scratch (SPaRCNet \cite{jing2023development}, ContraWR \cite{yang2021self}, FFCL \cite{li2022motor}, CNN-Trans \cite{peh2022transformer}, and ST-Trans \cite{song2021transformer}) and EEG FMs (MMM \cite{yi2023learning}, BENDR \cite{kostas2021bendr}, BIOT \cite{yang2023biot}, EEGPT \cite{wang2024eegpt}, LaBraM \cite{jiang2024large}, CBraMod \cite{wang2025cbramod}, Gram \cite{li2025gram}, NeuroLM~\cite{jiang2025neurolm}{, REVE~\cite{el2026reve}, NeurIPT~\cite{fang2026neuript}, CodeBrain~\cite{ma2025codebrain} and Uni-NTFM~\cite{chen2025uni}}).

\section{Results \& Discussion}
\begin{figure}
    \centering
    \includegraphics[width=1\linewidth]{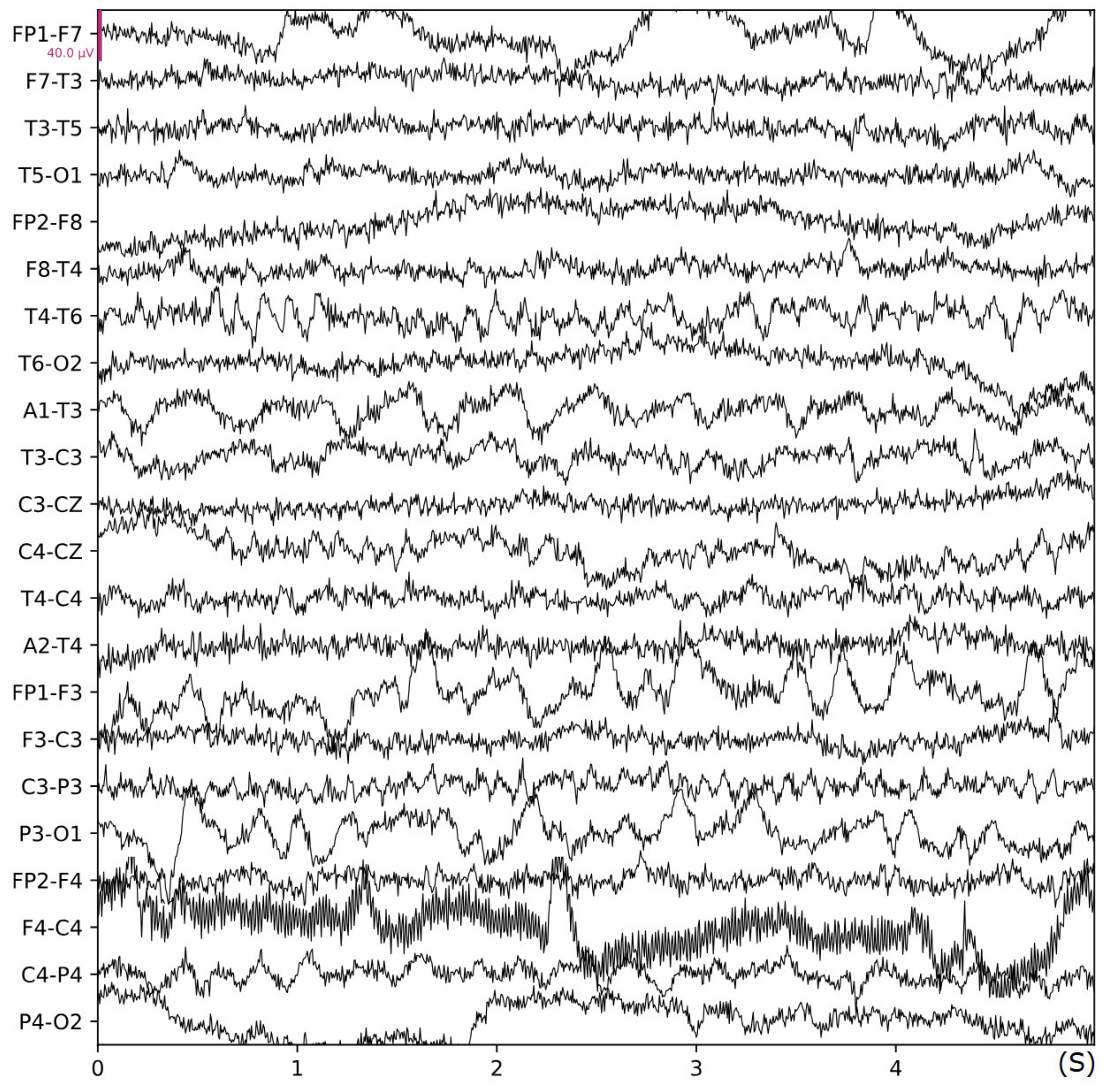}
    \caption{{Visualization of SSMDP-generated EEG traces.}}
    \label{fig:gen}
\end{figure}

We evaluated our model on the multi-event TUEV EEG classification task and compared its performance against several strong baselines, including those trained from scratch, EEG FMs that involved a pretraining stage using multiple datasets. In addition, we conducted ablation studies to investigate the contributions of key components in EEGDM, such as input noising strategies, latent activity extraction methods, the hierarchical structure captured in the learned latent representations, latent fusion mechanisms, {diffusion pretraining, and hyperparameter sensitivity}. Moreover, we examined the model’s ability to generalize to {other EEG datasets (CHB-MIT \cite{guttag2010chb} and IIIC\cite{jing2023development, ge2021deep})} as well as EEG data sampled from different frequencies. A visualization of EEG generated by SSMDP {can be found in Fig.~\ref{fig:gen}.}

\subsection{Main Results}
In Table~\ref{table:main-res}, we report the mean and standard deviation of ``kappa'', ``BAcc'', and ``WF1'', averaged over five runs with different random seeds. {The baselines are categorized according to their training strategies: \textit{Trained from scratch} indicates no pretraining is used; \textit{finetuning of pretrained parameters} indicates that pretrained weights are further updated for downstream tasks; and \textit{Freezing of pretrained parameters} indicates that pretrained weights are kept fixed during downstream evaluation. We further distinguish whether pretraining is conducted on a single dataset (non-FM) or on multiple heterogeneous datasets (FM)}. Detailed model sizes were also provided. In general, FMs exhibited a scaling law, with performance improving as model size increased. Furthermore, FMs that finetuned pretrained parameters typically outperformed those that froze them for efficiency. However, freezing pretrained parameters is often preferable due to lower training costs and improved efficiency, while still demonstrating the ability to represent richer and more universal EEG features. Our method achieved performance comparable to, and in most cases exceeding, that of the second-best baseline, Gram-L, while being approximately 19× more lightweight.

\subsection{Ablations}
We examined the impacts of each component, reporting the mean and standard deviation of ``kappa'', ``BAcc'', and ``WF1''.

\subsubsection{Signal Embedding and Noise Schedule of SSMDP}

Table \ref{table:alb-misc} shows the impact of $\mu$-law companding, highlighting the importance of handling the wide dynamic range of EEG signals. {In the TUEV dataset, this companding is applied to less than 2.5\% of the data (1.47\%, 2.38\%, and 2.03\% for the training, validation, and test sets, respectively), yet it effectively constrains the signal range to [-2, 2] and reduces the standard deviation by 5--10$\times$ to approximately 0.25.} In Section \ref{para:noise-sch}, we discuss the necessity of using a cosine noise schedule in the forward diffusion process. Here, we compare it against a diffusion model trained with the standard linear noise schedule \cite{ho2020denoising}, followed by training a classifier on the learned latent representations. As shown in Table \ref{table:alb-misc}, the linear noise schedule degrades classification performance. Therefore, we adopt the cosine noise schedule.

\subsubsection{Latent Activities and Representations of SSMDP}\label{sec:latentSSMDP}
When extracting latent activities from SSMDP, we found that minimizing the effect of the diffusion forward process, by using a noiseless forward process and a small diffusion step, enhanced the quality of latent activities and consequently improved performance, as shown in Table \ref{table:alb-misc} and Table \ref{table:alb-diff-step}. {This confirms that SSMDP representation captures informative latent structures intrinsically, without relying on noise injection or iterative denoising.}

We also analyzed the performance of different latent activities of SSMDP and various pooling strategies. As shown in Table \ref{table:alb-latent-type-pool}, when the pooling strategy is fixed, models using the latent activities of the gate channels outperformed those using other activities. Interestingly, standard deviation pooling, although rarely reported in the literature, outperformed average pooling, presumably because it better captured variability in time-series data.

\subsubsection{Ablation of LFM}
While the convolution kernels of the bidirectional SSMDP captured global context (see \eqref{eq:ssm-conv}), it is worth investigating whether the layers in SSMDP still exhibit a hierarchical structure, where shallow layers capture local patterns and deep layers capture global, high-level patterns \cite{gu2021efficiently}. To examine the internal hierarchy of the proposed SSMDP, we trained LFM using only a fraction of the SSMDP's layers. Since this reduced the size of the LFM-CL in LFM, the depth of the LFM-T was increased to compensate for the smaller model size. As shown in Table \ref{table:alb-half}, evaluation metrics generally improved when deeper layers were used, indicating that latent activities from deeper layers are more effective than those from earlier layers. Furthermore, incorporating latent activities from earlier layers in addition to deeper ones further improved performance.

We also explored two alternative methods with no trainable parameters: ``no fusion'', where the EEG channel ($C$), layer ($n$), and pool ($p$) axes of latent tokens were simply flattened, and ``mean fusion'', where latent fusion was performed by averaging along the layer ($n$) dimension. Since replacing LFM-CL removed a substantial part of the LFM, we increased the depth of the LFM-T to match the total number of trainable parameters in the base model for a fair ablation. The results in Table \ref{table:alb-fusion} underscore the importance of our latent fusion.

\begin{table}[!t]
\setlength{\tabcolsep}{3pt}
\centering
\caption{{Generalization (\%) of SSMDP pretrained on TUEV (with interictal but no {labeled ictal} seizures) to Cross Task \& Domain Downstream CHB-MIT (with pediatric seizures)}}

\label{table:chb}
\begin{tabular}{l r r r r}
\hline
\textbf{Model} & \multicolumn{1}{c}{\textbf{Size}} & \multicolumn{1}{c}{\textbf{BAcc}} & \multicolumn{1}{c}{\textbf{AUC-PR}} & \multicolumn{1}{c}{\textbf{AUROC}} \\
\hline
\multicolumn{5}{c}{\textbf{\textit{Training from Scratch on CHB-MIT (non-FM)}}}\\
SPaRCNet\cite{jing2023development}&0.79M& 58.76 ± 1.91 & 12.47 ± 1.19 & 81.43 ± 1.48\\
ContraWR\cite{yang2021self} &1.6M&63.44 ± 0.02& 22.64 ± 1.74 & 80.97 ± 1.14\\
FFCL\cite{li2022motor}   &    2.4M & 62.62 ± 1.04 & 20.49 ± 3.46 & 82.71 ± 0.51\\
CNN-Trans\cite{peh2022transformer}&3.2M&63.89 ± 0.67 & 24.79 ± 2.27 & 86.62 ± 0.82\\
ST-Trans\cite{song2021transformer} &3.5M & 59.15 ± 1.95 & 14.22 ± 0.94 & 82.37 ± 4.91\\
\hline
\multicolumn{5}{c}{\textbf{\textit{Pretrained with {Ictal} Seizure Dataset (FM)}}}\\

BIOT\cite{yang2023biot}&3.2M&70.68 ± 4.57 & 32.77 ± 4.60 & 87.61 ± 2.84\\
LaBraM-Base\cite{jiang2024large}&5.8M& 70.75 ± 3.58&32.87 ± 4.02&86.79 ± 1.99\\
CBraMod~\cite{wang2025cbramod}& 4.0M& {73.98 ± 2.84}
& {36.89 ± 3.82}
&{88.92 ± 1.54}
\\
{EEGMamba~\cite{wang2025eegmamba}} & {3.3M} & {\underline{74.45 ± 2.46}} & {38.85 ± 4.18} & {\underline{89.38 ± 1.61}}\\

{CodeBrain~\cite{ma2025codebrain}} & {15.17M}  & {72.73 ± 2.40} & {\underline{43.77 ± 2.88}} & {\textbf{89.61 ± 1.74}} \\

\hline
\multicolumn{5}{c}{\textbf{\textit{Pretrained without {Ictal} Seizures Data (non-FM)}}}\\
\textbf{EEGDM (Ours)} & 12.8M & \textbf{85.82 ± 1.04} & \textbf{48.45 ± 0.48} & {{89.26 ± 0.86}}
\\
\hline
\end{tabular}
\end{table}

\begin{table}[!t]
\caption{Generalization (\%) EEGDM to different sampling rates (TUEV).}
\label{table:aux-sampling}
\centering
\begin{tabular}{l l r r r}
\hline
\textbf{Train} & \multicolumn{1}{c}{\textbf{Test}} & \multicolumn{1}{c}{\textbf{Kappa}} & \multicolumn{1}{c}{\textbf{BAcc}} & \multicolumn{1}{c}{\textbf{WF1}} \\
\hline
\textbf{200 Hz} & \textbf{200 Hz} & \textbf{74.23 ± 1.36} & \textbf{75.57 ± 1.47} & \textbf{86.88 ± 0.66}\\
200 Hz & 190 Hz & \underline{70.26 ± 2.00} & \underline{72.78 ± 1.38} & \underline{85.17 ± 1.02}\\
200 Hz & 180 Hz & {66.54 ± 1.60} & {68.36 ± 2.07} & {83.38 ± 0.81}\\
\hline
\end{tabular}
\end{table}

\subsubsection{{Ablation of Diffusion Pretraining}}

{
Under the same backbone and hyperparameters of SSMDP and LFM, we replace the diffusion pretraining objective with a masked reconstruction objective. Specifically, we implement masked reconstruction, where 50\% of the EEG signal is masked using five non-overlapping temporal segments, and the model is trained to reconstruct the missing portions by minimizing the MSE loss. After pretraining, the subsequent finetuning procedure remains unchanged. As shown in Table~\ref{tab:alb-mask}, the diffusion-pretrained model significantly outperforms the masked reconstruction ablation, showing evidence that the performance gains are not attributed solely to the SSMDP architecture. Instead, our diffusion pretraining itself plays a critical role in improving the quality, robustness, and transferability of the learned EEG representations.}

\subsubsection{{Bidirectional SSM}}
{To isolate the effect of bidirectional modeling, we replace the bidirectional S4D modules with capacity-matched unidirectional counterparts on TUEV. As shown in Table~\ref{table:alb-misc}, bidirectional modeling provides a clear performance advantage. We therefore adopt bidirectional S4D across all experiments.}

\begin{table}[]

        \centering
        \caption{{Ablation of Diffusion Pretraining on TUEV (\%).}}
        
        \begin{tabular}{crrr}
        \hline
            {\textbf{Pretraining}} & \multicolumn{1}{c}{{\textbf{Kappa}}} & \multicolumn{1}{c}{{\textbf{BAcc}}} & \multicolumn{1}{c}{{\textbf{WF1}}} \\
        \hline
             {\textbf{Diffusion Pretraining}} & {\textbf{74.23 ± 1.36}} & {\textbf{75.57 ± 1.47}} & {\textbf{86.88 ± 0.66}} \\
             {Masked Reconstruction} & {\underline{59.69 ± 2.04}} & {\underline{54.98 ± 2.54}} & {\underline{78.81 ± 1.24}}\\
        \hline
        \end{tabular}
    \label{tab:alb-mask}
\end{table}

\subsection{Additional Analyses}

\subsubsection{Generalization Beyond the Pretrained Dataset}\label{sec:chb-mit}
We investigated whether our SSMDP, pretrained on the TUEV dataset {(containing interictal but no labeled ictal} seizures), could generalize to {two cross-task and cross-domain downstream datasets: CHB-MIT and IIIC, which differ in terms of task definition (IED detection vs. seizure detection and seizure type classification), the number of EEG channels (22 vs. 16 and 16), segment duration (5s vs. 10s and 10s), and the presence of class imbalance}.

Our method inherently handles any number of channels; hence, SSMDP can be directly applied to the first and second 5-second segments of CHB-MIT and IIIC. We compared our performance with FMs previously evaluated on the same dataset, as summarized in Table~\ref{table:chb}. This comparison followed the experimental setup of \cite{wang2025cbramod} and utilized BAcc, the area under the precision–recall curve (AUC-PR), and the area under the receiver operating characteristic curve (AUROC) as evaluation metrics. While FMs generally outperform models trained from scratch, it is noteworthy that existing FMs have been pretrained on seizure EEG data from either the same or different datasets. In contrast, our method, pretrained only on TUEV, demonstrated strong generalization, achieving superior performance against EEG FMs across most metrics.

{While CHB-MIT assesses the generalization capability for binary classification, IIIC dataset~\cite{jing2023development, ge2021deep} evaluated EEGDM on multi-class seizure subtype classification. We compare it against strong EEG FM baselines, following their experimental settings and hyperparameters: BIOT~\cite{yang2023biot}, CBraMod~\cite{wang2025cbramod}, CodeBrain~\cite{ma2025codebrain}, EEGMamba~\cite{wang2025eegmamba}, and Gram~\cite{li2025gram}. Table~\ref{tab:iiic} reports the results across five random seeds. To assess robustness against the strongest baseline (Gram~\cite{li2025gram}), we conducted a 30-seed evaluation using paired t-tests and Wilcoxon signed-rank tests. These tested the hypothesis that EEGDM outperforms Gram across BAcc, Kappa, and WF1. EEGDM achieved consistently strong, statistically significant improvements, demonstrating robust generalization.}

\subsubsection{EEGDM Label Efficiency}
In addition to cross-domain generalization, we also studied the generalization ability of the proposed method under the constraint of data scarcity. Following the same methodology as described in section \ref{sec:chb-mit}, we reduced the amount of training data to 10\%, 5\%, and 1\% of their original size, and reported the performance of the trained model on the same test data. As shown in Table \ref{table:data-scarce}, EEGDM remains competitive and label-efficient.

\begin{table}[!t]
\caption{Generalization of EEGDM to CHB-MIT dataset and its label efficiency (\%).}
\label{table:data-scarce}
\centering
\begin{tabular}{l r r r r}
\hline
\textbf{Model} & \multicolumn{1}{c}{\textbf{BAcc}} & \multicolumn{1}{c}{\textbf{AUC-PR}} & \multicolumn{1}{c}{\textbf{AUROC}} \\
\hline
\multicolumn{4}{c}{\textbf{\textit{Finetuned with 100\% Training Data}}}\\
BIOT\cite{yang2023biot}&70.68 ± 4.57 & 32.77 ± 4.60 & 87.61 ± 2.84\\
LaBraM-Base\cite{jiang2024large}& 70.75 ± 3.58&32.87 ± 4.02&86.79 ± 1.99\\
CBraMod\cite{wang2025cbramod}& {73.98 ± 2.84}&{36.89 ± 3.82}&\underline{88.92 ± 1.54}\\
\hline
\multicolumn{4}{c}{\textbf{\textit{Finetuned on smaller subsets of Training Data}}}\\
\textbf{EEGDM (Ours)} & \textbf{85.82 ± 1.04} & \textbf{48.45 ± 0.48} & \textbf{89.26 ± 0.86}\\
\textbf{EEGDM (Ours) 10\%} & \underline{81.97 ± 4.76} & \underline{43.93 ± 4.63} & {88.11 ± 0.49}\\
\textbf{EEGDM (Ours) 5\%} & {80.93 ± 3.47} & {39.03 ± 3.98} & 88.14 ± 0.60\\
\textbf{EEGDM (Ours) 1\%} & 69.56 ± 16.0 & 37.87 ± 1.57 & 87.72 ± 0.82\\

\hline
\end{tabular}
\end{table}

\subsubsection{Generalization to Different Sampling Rates}
EEGDM preserved the time-continuous form of SSMs, enabling it to process EEG signals at different sampling rates. Table \ref{table:aux-sampling} presents the performance of models trained on data with a 200 Hz sampling rate and tested on data with different sampling rates using TUEV. The results showed that performance degraded slightly but largely remained intact. Nevertheless, this capability is unique to the SSM-based method and not directly applicable to other EEG FMs.

\begin{table}[]
\setlength{\tabcolsep}{2pt}
\caption{{Comparative results (\%) on the IIIC seizure classification dataset. Statistical tests are conducted between the top two methods. 
$^\ast / ^{\ast\ast}$ and $\dagger / \dagger\dagger$ denote statistical significance at $p < 0.01$ and $p < 0.001$ for the Wilcoxon signed-rank test and paired $t$-test, respectively.}}\label{tab:iiic}
\centering
\begin{tabular}{l rl rl rl}
\hline
{\textbf{Model}} & \multicolumn{2}{c}{{\textbf{Kappa}}} & \multicolumn{2}{c}{{\textbf{BAcc}}} & \multicolumn{2}{c}{{\textbf{WF1}}} \\
\hline

\hline
\multicolumn{7}{c}{{\textbf{\textit{5 Seeds (Seed 0$\sim$4)}}}}\\
{BIOT~\cite{yang2023biot}}  & {36.29 ± 1.75} & 
& {49.06 ± 1.44} &
& {48.18 ± 1.49} & 
\\
{CBraMod~\cite{wang2025cbramod}} & {40.77 ± 0.25} & 
& {49.33 ± 0.29} & 
& {51.02 ± 0.30} &
\\
{CodeBrain~\cite{ma2025codebrain}} & {28.25 ± 0.70} &
& {38.74 ± 1.35} &
& {40.28 ± 0.64} &
\\
{EEGMamba~\cite{wang2025eegmamba}} & {44.03 ± 0.14} & 
& {52.84 ± 0.23} &
& {54.64 ± 0.13} &
\\
{Gram~\cite{li2025gram}} & {\underline{50.48 ± 2.21}} &  & {\underline{59.37 ± 2.16}} &  & {\underline{60.20 ± 1.74}} &  \\
{\textbf{EEGDM (Ours)}} & {\textbf{52.18 ± 0.67}} & & {\textbf{59.56 ± 0.84}} & & {\textbf{61.53 ± 0.62}} &\\

\hline
\multicolumn{7}{c}{{\textbf{\textit{30 Seeds (Seed 0$\sim$29)}}}}\\

{Gram~\cite{li2025gram}} & {50.27 ± 1.56} & {$\dagger\dagger$**}& {58.89 ± 1.53} & {$\dagger$*}& {59.97 ± 1.31} & {$\dagger\dagger$**}\\

{\textbf{EEGDM (Ours)}} & {\textbf{52.24 ± 0.60}} & & {\textbf{59.75 ± 0.67}} & & {\textbf{61.52 ± 0.57}}& \\

\hline
\end{tabular}

\end{table}

\subsubsection{{Hyperparameter Sensitivity}}

{
The hyperparameters of LFM, specifically the number of fusion tokens $N$ and the depth of the LFM-T, were selected based on empirical studies on TUEV using a small subset of the training and validation data. We further analyze the sensitivity of LFM with respect to these parameters. As shown in Table~\ref{tab:alb-hyp}, the model is relatively stable across a wide range of fusion token numbers. Performance degrades when $N$ is too small, likely due to insufficient cross-layer information aggregation, and when $N$ is excessively large, potentially due to increased redundancy and optimization difficulty. Similarly, increasing the depth of the LFM-T yields diminishing returns, with deeper configurations showing signs of mild overfitting and increased variance beyond a certain depth, while overall performance remains stable.
}

\begin{table}[]
        \setlength{\tabcolsep}{3pt}
        \centering
        \caption{{Hyperparameter sensitivity analysis of the LFM (\%).\\ * indicate the default value.}}
        \label{tab:alb-hyp}
        
        \begin{tabular}{c|c|rrr}
        \hline
            {\textbf{Component}} & {\textbf{Value}} & \multicolumn{1}{c}{{\textbf{Kappa}}} & \multicolumn{1}{c}{{\textbf{BAcc}}} & \multicolumn{1}{c}{{\textbf{WF1}}} \\
        \hline
             & {8} & {71.40 ± 1.16} & {74.23 ± 1.80} & {85.60 ± 0.63}\\
            {\# of fusion tokens} & {16*} & {\textbf{74.23 ± 1.36}} & {\textbf{75.57 ± 1.47}} & {\textbf{86.88 ± 0.66}} \\
             & {24} & {71.52 ± 1.65} & {73.03 ± 2.55} & {85.55 ± 0.85}\\
        \hline
            & {4} & {70.74 ± 1.68} & {73.05 ± 1.93} & {85.04 ± 0.91} \\
            {\# of encoder blocks} & {8*} & {\textbf{74.23 ± 1.36}} & {\textbf{75.57 ± 1.47}} & {\textbf{86.88 ± 0.66}} \\
            & {12} & {72.71 ± 1.52} & {74.40 ± 1.17} & {86.20 ± 0.75} \\
        \hline
        \end{tabular}
\end{table}

\subsubsection{{Performance Analysis}}
{The computational requirements of the best-performing methods are summarized as tuples in the form of (pretraining data size, GPU configuration, training duration): EEGDM (11.37 hours EEG, 1$\times$ RTX 4090 24GB, 23 hours), Gram-L (7,000 hours EEG, 8$\times$ RTX A800, 13 days), EEGMamba (16,000 hours EEG, 1$\times$ RTX A5000, 5 days), and CodeBrain (9,000 hours EEG, 6$\times$ A100 for 10 hours followed by 2$\times$ A100 for 24 hours). EEGDM achieves competitive performance while requiring substantially fewer pretraining data and computational resources.} {During fine-tuning, EEGDM latent tokens from all layers were precomputed to eliminate redundant SSMDP forward passes. Although this incurs an additional storage overhead of approximately $4.7\times$ the original data size, it remains manageable with standard data-loading pipelines and provides approximately $6.78\times$ acceleration in training. To evaluate inference latency, we simulate real-time deployment (batch size of 1, without pre-computation), achieving a throughput of 33.25 samples per second with 1.5 GB GPU memory usage. The average latency of 30.08 ms per sample indicates that the model is suitable for low-latency inference for clinical scenarios.}

\subsubsection{{Complexity Analysis}}

{
Given an EEG input $\mathbf{X} \in \mathbb{R}^{C \times L}$ with $C$ channels and $L$ samples, each channel is divided into $P$ temporal patches. The SSMDP backbone has linear complexity $\mathcal{O}(CP)$. The LFM-CL uses a fixed number of fusion tokens $N$, which is independent of the input length $L$ and the number of channels $C$. This makes the fusion stage efficient, with complexity $\mathcal{O}(CP)$. Within LFM, the LFM-T applies self-attention over $N\cdot P$ tokens, giving complexity $\mathcal{O}(P^2)$. In contrast, standard EEG FMs apply self-attention over all channel-patch tokens, resulting in quadratic complexity $\mathcal{O}((CP)^2)=\mathcal{O}(C^2P^2)$. 
}

\section{Conclusion and Future Work}

This work presented EEGDM, a diffusion-based framework for self-supervised EEG representation learning. EEGDM comprises SSMDP pretrained within the DDPM framework to learn and generate high-granularity EEG. The learned representations are subsequently aggregated by LFM for downstream classification. Empirically, EEGDM {is} {competitive with, in most cases exceeding} state-of-the-art methods on EEG event classification tasks when assessed using the TUEV dataset, despite having significantly fewer training samples and being approximately 19× smaller than current EEG FMs. Moreover, SSMDP demonstrates strong cross-task and cross-domain generalization (i.e., CHB-MIT and IIIC), despite requiring neither large-scale heterogeneous EEG corpora for pretraining nor extensive downstream labels for adaptation. Comprehensive ablation studies validated the effectiveness of each EEGDM component. By achieving robust state-of-the-art performance with strong data and label efficiency, our method demonstrates promising clinical relevance for seizure diagnosis and management, particularly in data-scarce clinical settings.

While our results show that EEGDM can learn strong representations without the large-scale pretraining typically required by FMs, these findings point to a promising direction for diffusion-based EEG FMs. Future work may further explore: (i) improving model efficiency to facilitate clinical deployment under limited computational resources; (ii) scaling diffusion-based representation learning through large-scale, multi-dataset EEG pretraining, potentially with redesigned conditioning mechanisms to accommodate heterogeneous recording settings; and (iii) examining the learned representations across a broader range of EEG paradigms and downstream beyond epilepsy-related tasks.

\section*{Acknowledgment}
The work was supported by the Ministry of Higher Education Malaysia through the Fundamental Research Grant Scheme (FRGS/1/2023/ICT02/XMU/02/1), and Xiamen University Malaysia through Xiamen University Malaysia Research Fund (XMUMRF/2022-C10/IECE/0039, XMUMRF/2024-C13/IECE/0049).

\bibliographystyle{IEEEtran}

\bibliography{bibliography}

\vfill
\end{document}